\documentclass[letterpaper, 10 pt, conference]{ieeeconf}  

\IEEEoverridecommandlockouts                              

\usepackage{times}
\usepackage{soul}
\usepackage{url}
\usepackage[hidelinks]{hyperref}
\usepackage[utf8]{inputenc}
\usepackage[small]{caption}
\usepackage{graphicx}
\usepackage{amsmath}
\usepackage{booktabs}
\usepackage{placeins}
\usepackage{amsmath,amsfonts,bm}

\def\eqref#1{equation~\ref{#1}}

\def\1{\bm{1}}

\DeclareMathAlphabet{\mathsfit}{\encodingdefault}{\sfdefault}{m}{sl}
\SetMathAlphabet{\mathsfit}{bold}{\encodingdefault}{\sfdefault}{bx}{n}

\usepackage{graphicx} 
\usepackage{algorithm}
\usepackage[noend]{algorithmic}
\usepackage{todonotes}
\usepackage{comment}
\usepackage{paralist}
\usepackage{subfig}
\usepackage{wrapfig}
\usepackage{bbm}
\usepackage{xspace}
\usepackage{bbold}

\usepackage{newfloat}
\usepackage{listings}

\usepackage{amstext}
\usepackage{amsmath}
\usepackage{amssymb}

\algsetup{indent=0.75em}
\algsetup{linenosize=\scriptsize}
\newcommand{\refinepartition}[0]{Refine}

\title{\LARGE \bf
Goal Space Abstraction in Hierarchical Reinforcement Learning via Set-Based Reachability Analysis}

\newcommand{\todoSM}[1]{}
\newcommand{\todoNM}[1]{}
\newcommand{\todoMZ}[1]{}

\newcommand{\states}{\ensuremath{\mathcal{S}}}
\newcommand{\partitions}{\ensuremath{\mathcal{G}}}
\newcommand{\partition}{\ensuremath{G}}
\newcommand{\edges}{\ensuremath{\mathcal{E}}}
\newcommand{\fwdmodel}{\ensuremath{\mathcal{F}_k}}

\newcommand{\policy}{\ensuremath{\pi}}
\newcommand{\highpolicy}{\ensuremath{\policy_{\text{High}}}}
\newcommand{\lowpolicy}[2]{\ensuremath{\policy_{\text{Low}}({#1},{#2})}}
\newcommand{\optlowpolicy}[2]{\ensuremath{\policy^*_{\text{Low}}({#1},{#2})}}
\newcommand{\reachrel}{\ensuremath{R_k}}
\newcommand{\env}{\ensuremath{E}}

\newcommand{\reachpart}{\ensuremath{\partitions_{R}}}
\newcommand{\unreachpart}{\ensuremath{\partitions_{U}}}
\newcommand{\ratio}{\ensuremath{r}}
\newcommand{\reachthreshold}{\ensuremath{t_{\text{re}}}}
\newcommand{\unreachthreshold}{\ensuremath{t_{\text{unre}}}}

\newcommand{\acronym}{GARA\xspace}

\begin{document}

%
%

\author{%
Mehdi Zadem$^1$,
Sergio Mover$^1$,
Sao Mai Nguyen$^{2,3}$
\thanks{$^1$LIX, CNRS, \'Ecole Polytechnique, Institut Polytechnique de Paris, zadem@lix.polytechnique.fr,
sergio.mover@lix.polytechnique.fr,}
\thanks{$^2$Flowers Team, U2IS, ENSTA Paris, Institut Polytechnique de Paris \& Inria, $^3$IMT Atlantique, Lab-STICC, nguyensmai@gmail.com,}
}

\maketitle

\begin{abstract}
Open-ended learning benefits immensely from the use of symbolic methods for goal representation as they offer ways to structure knowledge for efficient and transferable learning.
However, the existing Hierarchical Reinforcement Learning (HRL) approaches relying on symbolic reasoning are often limited as they require a manual goal representation.
The challenge in autonomously discovering a symbolic goal representation is that it must preserve critical information, such as the environment dynamics. 
In this paper, we propose a developmental mechanism for goal discovery via an emergent representation that abstracts (i.e., groups together) sets of environment states that have similar roles in the task.
%
We introduce a Feudal HRL algorithm that concurrently learns both the goal representation and a hierarchical policy.
The algorithm uses symbolic reachability analysis for  neural networks to approximate the transition relation among sets of states and to refine the goal representation.
We evaluate our approach on complex navigation tasks, showing the learned representation is interpretable, transferrable and results in data efficient learning.
\end{abstract}

\footnote{This paper has been accepted for publication in IEEE International Conference on Development and Learning (ICDL) 2023
IEEE Copyright Notice:
© 2023 IEEE. Personal use of this material is permitted. Permission from IEEE must be obtained for all other uses, in any current or future media, including reprinting/republishing this material for advertising or promotional purposes, creating new collective works, for resale or redistribution to servers or lists, or reuse of any copyrighted component of this work in other works.}

\section{Introduction}


Symbol emergence is key for developmental learning to tackle the curse of dimensionality and scale up to  open-ended high-dimensional sensorimotor space, by allowing symbolic reasoning, compositionality, hierarchical organisation of the knowledge, etc. While symbol emergence has been recently investigated for the sensor data, action symbolization can lead to a repertoire of various movement patterns by bottom-up processes, which can be used by top-down processes such as compostion to form an action sequence, planning and reasoning for more efficient learning, as reviewed in \cite{Taniguchi2018ITCDS}. Sensorimotor symbol emergence thus is key to scaling up low-level actions into complex actions for open-ended learning, using compositionality and hierarchy.

Action hierarchies are the core idea of Hierarchical Reinforcement Learning (HRL) \cite{Barto2003DEDS} that decomposes a task into easier subtasks. In particular, in Feudal HRL \cite{feudal} a high-level agent selects subgoals that a low-level agent learns to achieve. The performance of Feudal HRL depends on the "hierarchical division of the available state space" \cite{feudal}, the representation of the goals that the high level agent uses to decompose a task.
%
Yet, only few algorithms learn it automatically \cite{DBLP:journals/corr/VezhnevetsOSHJS17}, while others either use directly the state space~\cite{nachum2018dataefficient} or manually provide a representation~\cite{hdqn,DBLP:conf/nips/ZhangG0H020}. The issue of symbol emergence within hierarchical reinforcement learning has little been investigated.

So far there is "no computational model that can reproduce the dynamics of symbol emergence in the real world itself" \cite{Taniguchi2018ITCDS}. Learning online an abstraction of the sensorimotor space grounded by the experienced data is all the more challenging if the abstraction needs to be amenable for transfer learning  in a new environment and to be interpretable.

%
%


In this paper, we tackle the problem of learning automatically, while learning the policy,
 a discrete goal representation from continuous observations 
that expresses the task structure for \emph{data-efficiency}, is \emph{transferrable}, and is \emph{interpretable}.

We introduce a 
  feudal HRL algorithm, \acronym (Goal Abstraction via Reachability Analysis), that develops a novel symbolic goal space representation while simultaneously learning a hierarchical policy from exploration data. The representation emerges through a developmental process, gradually gaining precision from a bottom-up manner, 
by leveraging data acquired from exploration. 
This discretisation of the environment is used to orient  top-down process of the goal-directed exploration, that in turn helps improving policies and this representation. We thus propose  a symbolic representation of the goal space  emerging dynamically from  bottom-up and top-down processes. 


%
%

%
%
 This paper focuses on deterministic and reward-sparse environments with continuous state spaces, highlights the emergence of a structured representation, and shows its impact in terms of data efficiency, transferability and interpretability.
 We propose: 

\begin{enumerate}
\item A symbolic representation of the goal space (Section~\ref{sec:framework}), obtained grouping together sets of similar states according to the state reachability relation, that leads to the emergence of action abstractions that can be chained into action sequences:  
  two states are similar if a low-level policy reaches the same set of similar states. This representation is interpretable as a directed graph, where nodes describe goals and edges a reachability relation among them.
  To build such representation, the algorithm symbolically analyzes a neural network approximating the state reachability relation, similarly to reachability analysis used in formal verification~\cite{DBLP:journals/ftopt/LiuALSBK21}.
\item A two level HRL algorithm, \acronym that represent the bottom-up and top-down processes that discover and learn online the goal representation via set-based reachability analysis 
(Section~\ref{sec:methodology}).

\item An experimental evaluation (Section~\ref{sec:evaluation}) demonstrating that the abstract goal representation can be learned  in a data-efficient manner. 
The experiments further validate that the representation of the goals enables transfer learning and is interpretable.
\end{enumerate}

\section{Framework}
\label{sec:framework}
\subsection{Feudal Hierarchical Reinforcement Learning}
Reinforcement Learning (RL) algorithms learn to take actions in different states of an environment in order to complete a task. An agent is a Markov Decision Process $M = (\states, A, P, r)$, where $\states$ is a set of states, $A$ is a set of actions, $P$ is a probabilistic transition function, and $r$ is a reward function.
At each execution time $t$, the agent is in a state $s_t \in \states$ and chooses an action $a$ to execute with a policy $\pi$ (i.e., $a \sim \pi(s_t)$). After executing the action, the agent receives a reward $r$ from the environment.
RL algorithms learn a policy $\pi$ that maximizes the expected future reward.
%
%
In Hierarchical Reinforcement Learning (HRL), multiple agents operate at different levels of the problem. Higher level agents decompose the initial learning task into subtasks, while lower level agents learn a policy to achieve such sub-tasks, either by further decomposing them into smaller subtasks or by learning concrete actions for solving them.

We adopt the Feudal HRL~\cite{feudal} architecture, where the hierarchy is composed of two levels: a high-level agent $\mathcal{A}^\text{High}$ samples a goal $g_t$ from a goal space $\mathcal{G}$. The goal space $\mathcal{G}$ can be either the environment states itself (i.e., $\mathcal{G}=\states$) or an \emph{abstract goal representation}, where a function maps states in $\states$ to goals in $\mathcal{G}$.
The high-level agent selects a goal $g_t$ using a high-level policy $\highpolicy$ (i.e., $g_t \sim \highpolicy(s_t)$). The policy $\highpolicy$ is learned to maximize the expected environment "external" reward $r^{ext}$.
Then, the selected goal $g_t$ is communicated to a low-level agent that trains to learn a policy $\lowpolicy{s_t}{g_t}$ that reaches $g_t$ from the current state $s_t$.
In our framework, we assume the low-level policy is a Universal Value-Function Approximator (UVFA)~\cite{uvfa}, where the goal itself is as an additional parameter to the policy. In such way, the low-level policy learns different behaviours depending on the goal $g_t$.

\subsection{Goal Abstraction}
\label{sec:abstraction}

We propose a goal abstraction that discretizes the continuous environment state space $\states$ into a finite number of disjoint subsets.
Formally, the abstract goal space is a \emph{partition}  $\partitions = \{\partition_0, \dots , \partition_n\}$ of the state space $\states$, that is $\left(\bigcup_{\partition \in \partitions}{\partition}\right) = \states$ and for all $\partition,\partition' \in \partitions$, $\partition \cap \partition' = \emptyset$ if $\partition \neq \partition'$.
A key intuition is that the partition $\partitions$ depends on the behavior of the low-level agent and, in particular, on the \textit{reachability relation} $\reachrel(\partition') \subseteq \states \times \states$. $\reachrel$ is such that $(s,s') \in \reachrel(\partition')$ if the state $s$ reaches the state $s' \in \partition'$ after applying the optimal low-level policy $\optlowpolicy{s}{\partition'}$ for $k$-steps.
We extend the notation of reachability relation $\reachrel$ to sets of states $\partition$ to represent the set of states that $\partition$ reaches using $\reachrel(\partition')$, i.e., $\reachrel(\partition, \partition') = \{s' \in \states \mid \exists s \in \partition. (s,s') \in \reachrel(\partition') \}$.
The abstract goal space, the partition $\partitions$, satisfies the property:
\begin{align}
\label{eq:prop}
\forall ~ \partition, \partition' \in \partitions.\reachrel(\partition, \partition') \subseteq \partition'.
\end{align}
This property expresses that all the states in a subset $\partition \in \partitions$ have a similar behavior in the environment, reaching states that are also similar (i.e., a bisimulation relation among the environment's states, see, e.g.~\cite{DBLP:books/daglib/0020348}).
A subset $\partition$ is said \textit{stable} with respect to $\partition'$ if $\reachrel(\partition, \partition') \subseteq \partition'$.
The main advantage of such goal abstraction is to capture the behavior of the low-level agent, while still decomposing the state space. In fact, the high-level agent will ask the low-level agent to reach target subgoals $s' \in \partition'$ from an initial subgoal $s \in \partition$, which should be feasible with respect to the low-level policies.

\section{Methodology}
\label{sec:methodology}

\subsection{Algorithm Outline}
Goal Abstraction via Reachability Analysis (\acronym) (Algorithm \ref{alg:feudalhrl}) is a Feudal HRL algorithm that learns, simultaneously, the agent policies and the goal abstraction $\partitions$.
%
\acronym learns to solve a task specified with a set of initial and target states in an environment $\env$.
The algorithm starts with the initial goal set $\partitions = \{\states\}$ and then iterates for $n_{\text{eps}}$ episodes. 

In each episode, \acronym starts in an initial state $s_{t}$ in the set $\partition_s \in \partitions$ and then selects a target set $\partition_d$ with the high-level policy $\highpolicy$, which the algorithm learns with the environment's reward signal ($r_t^{\text{ext}}$). The low-level agent trains a UVFA low-level policy $\pi_{Low}$ for reaching the target set $\partition_d$ from the state $s_t$, using as reward a function ($\text{reward}_{\text{Low}}$) of the environment reward ($r_t^{\text{ext}}$) and the reward for reaching the target set $\partition_d$ 
($\mathbb{1}_{\partition_d}(s_{t+1})$).
\acronym uses Q-learning
for learning the high-level policy (the high-level agent's states are finite), and Deep Q-learning \cite{dqn} for the low-level policy.

The key insight of \acronym\ is to gradually learn a \emph{stable} goal representation $\partitions$ (see \eqref{eq:prop}), refining the start set $\partition_s$ of each abstraction edge $(\partition_s,\partition_d) \in \edges$ that the agent visited after each episode.
The algorithm uses the data acquired during training to approximate the \emph{unknown} reachability relation $\reachrel$.
\acronym uses the result of each episode, stored in the memory $\mathcal{D}$ (Line~\ref{line:updatememory}), to train the \textit{k-forward model} $\fwdmodel : \states \times \partitions \rightarrow \states$ (line~\ref{line:refinemodel}), a neural network that approximates the reachability relation $\reachrel$. Given a state $s_t$ and a target set $\partition_d$, $\fwdmodel(s_t, \partition_d)$ predicts the state $s_{t+k}$ that the agent reaches after applying the low-level policy $\lowpolicy{s_t}{\partition_d}$ for a sequence of $k$ steps.
In our implementation, $\fwdmodel$ is a fully connected neural networks and it uses a MSE loss function (see~\cite{garareport} for details).
\acronym then \textit{refines} the goal abstraction $\partitions$ with the \textit{\refinepartition} function (line~\ref{line:refinepartition}).

Observe that, during an episode, \acronym may either not reach the target goal set or the initial partition may have a single element (i.e., $\partitions = \{\states\}$, so any state in $\states$ reaches another state in $\states$). In such cases, the criteria for refining an abstract goal set $\partition$ is to separate the states in $\partition$ reached during the episodes from the other ones (using the memory $\mathcal{D}$).
\acronym uses a standard density-based clustering approach (DBSCAN \cite{10.5555/dbscan}) to identify a cluster of frequently visited states. Note that, in later episode the algorithm will eventually try to explore the new subset of $\partition$ containing the unreachable states.

\begin{algorithm}[tbh]
\caption{Goal Abstraction via Reachability Analysis}
\label{alg:feudalhrl}
\begin{algorithmic}[1]
\footnotesize
\REQUIRE Learning environment $\env$.
\ENSURE Computes the high-level and low-level policies $\highpolicy$ and $\pi_\text{Low}$
\STATE $\mathcal{D} \leftarrow \emptyset$, $\partitions \leftarrow \states$, $\text{episodes} \leftarrow 0$
\FOR{$\text{episode} \leq n_{\text{eps}}$}
  \STATE $\text{episode} \leftarrow \text{episode} + 1$, $t \leftarrow 0$, $\edges \leftarrow \emptyset$
  \STATE $s_{\text{init}} \leftarrow$ initial state from $\env$, $s_t \leftarrow s_{\text{init}}$
  \STATE $\partition_s \leftarrow $ $\partition \in \partitions$ such that $s_t \in \partition$
  \STATE $\partition_d$ $ \sim \highpolicy(\partition_s)$
  \WHILE{true}
    \STATE $\edges \leftarrow \edges \cup \{(\partition_s, \partition_d)\}$
    \STATE $a_t \sim \lowpolicy{s_t}{\partition_d}$
    \STATE $(s_{t+1}, r_t^{\text{ext}}, \text{done}) \leftarrow$ execute the action $a_t$ from $s_t$ in $\env$
    \STATE $r_t^\text{Low} = \text{reward}_{\text{Low}}
    (\mathbb{1}_{\partition_d}(s_{t+1}), 
    r_t^{\text{ext}})$
    \STATE Update $\pi_\text{Low}$ with reward $r_t^\text{Low}$ \label{line:dqn}
%
    \IF {not done}
      \STATE $s_t \leftarrow s_{t+1}$, $t \leftarrow t + 1$
      \IF{$t~mod~k = 0$ or $s_t \in \partition_d$}
        \STATE Add $(s_{\text{init}}, \partition_s, s_t, \partition_d, r_t^{\text{ext}}, \text{done})$ to $\mathcal{D}$
        \label{line:updatememory}
        \STATE Update $\highpolicy$ with reward $r_t^{\text{ext}}$ \label{line:qlearning}
        %
        \STATE $s_{\text{init}} \leftarrow s_t$
        \STATE $\partition_s \leftarrow $ $\partition \in \partitions$ such that $s_t \in \partition$
        \STATE $\partition_d$ $ \sim \highpolicy(\partition_s)$
      \ENDIF
    \ELSE
      \STATE break the \textbf{while} loop and terminate the current episode            
    \ENDIF
  \ENDWHILE 
   \IF{$\partitions \neq \states$ \AND $G_d$ is reached in $\mathcal{D}$}
     \STATE Update $\fwdmodel$ with the data from $\mathcal{D}$ \label{line:refinemodel}
     \STATE $\partitions \leftarrow \textit{\refinepartition}(\partitions,\edges,\fwdmodel)$
     \label{line:refinepartition}
   \ELSE
     \STATE $\partitions \leftarrow Cluster(\partitions)$
     \label{line:cluster}
   \ENDIF      
\ENDFOR 
\end{algorithmic}
\end{algorithm}

\subsection{Refining Goal Abstraction via Reachability Analysis}

\begin{algorithm}[h]
\caption{\textbf{\refinepartition}($\partitions$,$\edges$,$\fwdmodel$)}
\label{alg:refinepartitions}
\begin{algorithmic}[1]
\footnotesize
\REQUIRE goal abstraction $\partitions$, set of edges $\edges$, and k-forward model $\fwdmodel$
\ENSURE goal abstraction $\partitions'$ that refines $\partitions$
\STATE $\partitions' \leftarrow \partitions$
\FOR{ $(\partition_s,\partition_d) \in \edges$ } \label{alg:refinepartitions:loop}
  \IF{$error(F_k(s \in \partition_s,\partition_d))$ stable}
    \STATE $(\reachpart, \unreachpart) \leftarrow \textit{SplitSet}(\partition_s, \partition_d, \fwdmodel)$ \label{alg:refinepartitions:reach}
    \STATE $\partitions' \leftarrow (\partitions' \setminus \{\partition_s\}) \cup \reachpart \cup \unreachpart$
  \ENDIF
\ENDFOR
\RETURN $\partitions'$
\end{algorithmic}
\end{algorithm}
\begin{algorithm}[h]
\caption{\textbf{SplitSet}($\partition_s$,$\partition_d$,$\fwdmodel$)}
\label{alg:setpartition}
\begin{algorithmic}[1]
\footnotesize
\REQUIRE start set $\partition_s$, target set $\partition_d$, and k-forward model $\fwdmodel$
\ENSURE sets of states $\reachpart$ and $\unreachpart$
\STATE $\reachpart \leftarrow \emptyset$, $\unreachpart \leftarrow \emptyset$, $L \leftarrow \{ \partition_s \}$
\WHILE{ $L \neq \emptyset$ }
  \STATE $\text{remove an element } \partition \text{ from } L$
  \STATE $\partition_{R} \leftarrow \textit{FnnReach}(\fwdmodel, \partition)$ \label{line:fnnreach}
  \STATE $r \leftarrow \frac{\text{Volume}(\partition_{R} \cap \partition_d)}{\text{Volume}(\partition_{R})}$
  \STATE \textbf{if} $r > \reachthreshold$ \textbf{then} $\reachpart \leftarrow \reachpart \cup \{\partition\}$
  \STATE \textbf{else if} {$r < \unreachthreshold$} \textbf{then}  $\unreachpart \leftarrow \unreachpart \cup \{\partition\}$
  \STATE \textbf{else} \textbf{then} $(\partition',\partition'') \leftarrow \textit{Split}(\partition)$,%
    $L \leftarrow L \cup \{\partition',\partition''\}$
\ENDWHILE
\RETURN $(\reachpart,\unreachpart)$
\end{algorithmic}
\end{algorithm}

\paragraph{Refining a Partition} 
the \textit{\refinepartition} algorithm (Algorithm~\ref{alg:refinepartitions}) refines the partition of the abstraction according to ~\eqref{eq:prop}.
\textit{\refinepartition} takes as input the current abstraction $\partitions$, a set $\edges \subseteq \partitions \times \partitions$ of transitions, and the k-forward model $\fwdmodel$, and computes a new  abstraction $\partitions'$. $\partitions'$ is a refinement of the abstraction $\partitions$ (i.e., $\bigcup_{\partition \in \partitions}{\partition} = \bigcup_{\partition \in \partitions'}{\partition}$ and for all $\partition' \in \partitions'$ there exists a set $\partition \in \partitions$ such that $\partition' \subseteq \partition$).

A transition $(\partition_s, \partition_d)$  is in the set $\edges$ if, in the last episode, the agent reached a state $s_k \in \partition_d$ from a state $s_0 \in \partition_s$ executing the low-level policy $\lowpolicy{s_0}{\partition_d}$ for $k$ steps.
We perform reachability analysis after $F_k$'s error has stabilised for the considered part of the environment. Meaning the low-level policies remain relatively unchanged.
Note that, in the the past episode, Algorithm~\ref{alg:feudalhrl} only acquired new data for the transitions $(\partition_s,\partition_d) \in \edges$,
and Algorithm~\ref{alg:refinepartitions} splits each ``start'' set $\partition_s$ with the  \textit{SplitSet} function (line~\ref{alg:refinepartitions:reach}).
%
%
\textit{SplitSet} computes a set $\reachpart$ of reachable and a set $\unreachpart$ of unreachable sets such that $\reachpart = \{\partition \mid \partition \subseteq \partition_s, \reachrel(\partition) \subseteq \partition_d\}$, $\unreachpart = \{\partition \mid \partition \subseteq \partition_s, \reachrel(\partition') \subseteq \overline{\partition_d}\}$, where $\overline{\partition_d}$ is the complement of $\partition_d$, and
$\bigcup_{\partition \in \reachpart \cup \unreachpart} \partition = \partition_s$.
%
%
Then, the algorithm obtains the refinement $\partitions'$ replacing in $\partitions$ the set $\partition_s$ with the sets from $\reachpart$ and $\unreachpart$.
Since representing non-convex sets is computationally expensive, we restrict $\partition$ to be convex (our implementation uses  intervals~\cite{rival2020introduction}). 

While standard partition refinement algorithms (e.g.,~\cite{DBLP:journals/siamcomp/PaigeT87}) compute the coarsest partition with respect to a relation (e.g., $\reachrel$), \textit{\refinepartition} only splits the source set of each edge in
the set $\edges$.
%
Thus, the refined abstraction $\partitions'$ may not satisfy the goal abstraction criteria in all the sets
(e.g., suppose the algorithm splits $\partition_s$ in two subsets, $\partition_s'$ and $\partition_s''$, and that in $\partitions$ there is a set $\partition_{\text{pre}}$ that can reach $\partition_s$; while $\partition_{\text{pre}}$ may not be stable, \textit{\refinepartition} does not split it).
%
%
In practice, eagerly splitting a set is counterproductive since we approximate the  relation $\reachrel$ with $\fwdmodel$. 
If needed, \acronym will refine the set in a later episode.

\paragraph{Splitting a Single Set via Reachability Analysis}
\textit{SplitSet} (Algorithm~\ref{alg:setpartition}) computes the subsets of $\partition_s$ that can reach the set $\partition_d$ (i.e., a \emph{preimage} of $\partition_d$ with respect to the reachability relation $\reachrel$).
%
As discussed earlier, we approximate $\reachrel$ with the k-forward model $\fwdmodel$. 
While there exist several algorithms that compute (an approximation of) the set of reachable states from a set of states (i.e., the image of the function $\fwdmodel$, $\{s' \mid \exists s,\partition_d.\ s' = \fwdmodel(s,\partition_d)\}$), there are no algorithms that directly compute a preimage.
%
%
So, \textit{SplitSet} reduces the preimage computation to several image computations (\textit{FnnReach} in the algorithm, which we implement with the tool Ai2~\cite{DBLP:conf/sp/GehrMDTCV18}).

\textit{SplitSet} computes the preimage of $\partition_d$ iteratively.
First, the algorithm (line~\ref{line:fnnreach}) computes the set of reachable states $\partition_r = \textit{FnnReach}(\fwdmodel, \partition_s)$.
$\partition_r$ can partially overlap the destination set $\partition_d$ (Figure~\ref{fig:splitpartition:1}).
Since the computation of \textit{FnnReach} introduces over-approximations,
the algorithm uses an approximation error $\ratio$ for determining if $\partition_r$ is completely contained or outside $\partition_d$.
$\ratio$ is the ratio of the volume of $\partition_r \cap \partition_d$, and the volume of $\partition_r$ (\textit{Volume} is the volume of a set). The value of the error $\ratio$ is between 0 and 1: it is 0 when $\partition_r \subseteq \overline{\partition_d}$, and is 1 when $\partition_r \subseteq \partition_d$.
There are three cases:
\begin{inparaenum}[(i)]
\item the algorithm splits $\partition$ when $\partition$ \emph{mostly intersects} both $\partition_d$ and $\overline{\partition_d}$ (if $\unreachthreshold \le \ratio \le \reachthreshold$, Figure~\ref{fig:splitpartition:1}). In our implementation, the function \textit{Split} halves the interval $\partition$ along one of the existing dimensions, obtaining two sets $\partition',\partition''$ that the algorithm will split recursively.
\item The algorithm adds $\partition$ to the set $\reachpart$ when $\partition$ is \emph{mostly contained} in $\partition_d$ ($r > \reachthreshold$, Figure~\ref{fig:splitpartition:2}).
\item The algorithm adds $\partition$ to the set $\unreachpart$ when $\partition$ is \emph{mostly outside} $\partition_d$ (if $\ratio < \unreachthreshold$, Figure~\ref{fig:splitpartition:4}).
\end{inparaenum}

Thus, \textit{SplitSet} computes the subsets of $\partition_s$ that mostly reach $\partition_d$ ($\reachpart$) and that mostly not reach $\partition_d$ ($\unreachpart$).
The thresholds $\reachthreshold$ and $\unreachthreshold$ ($0 \le \unreachthreshold < \reachthreshold \le 1$) influence the precision of the analysis and the number of final sets: increasing $\reachthreshold$ will result in more precise, but more numerous, reachable sets.


\newcommand{\splitfigwidth}{0.15}

\begin{figure}[tbh]
  \begin{center}
    \subfloat[{\footnotesize$\unreachthreshold \le \ratio \le \reachthreshold$}\label{fig:splitpartition:1}]{%
      \includegraphics[width=\splitfigwidth\textwidth]{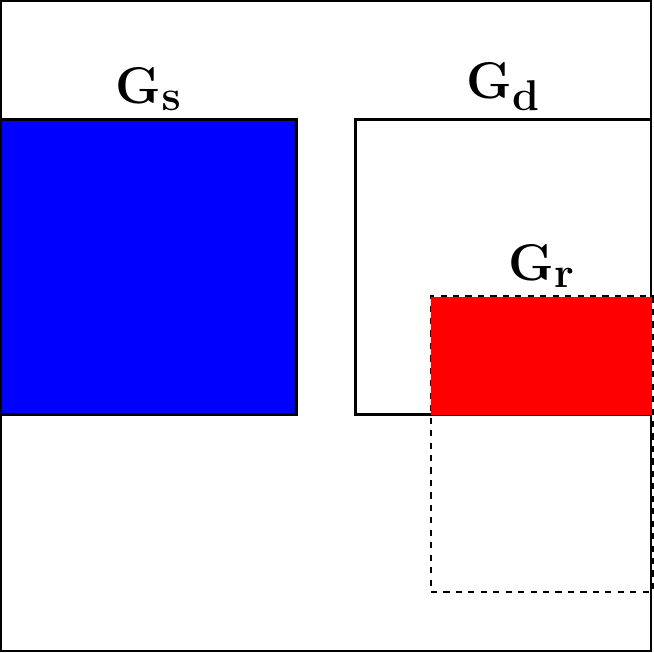}
    }
    \subfloat[{\footnotesize$\reachthreshold < \ratio$}\label{fig:splitpartition:2}]{%
      \includegraphics[width=\splitfigwidth\textwidth]{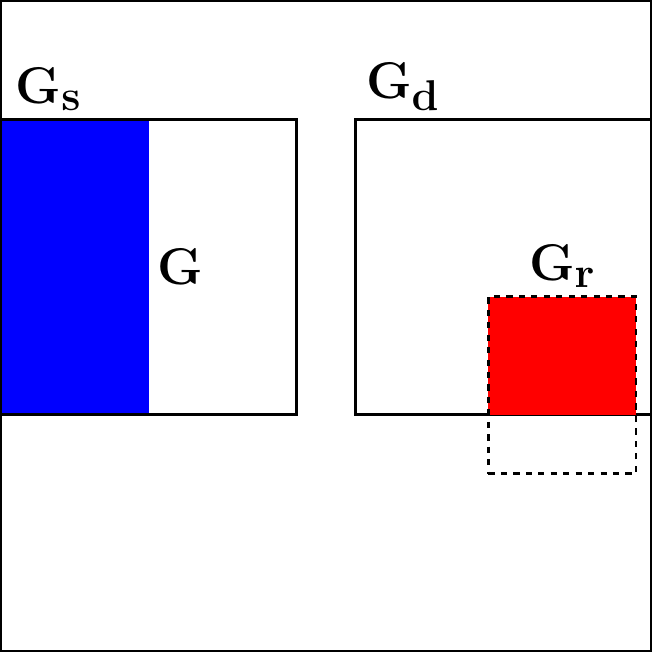}
    }
    \subfloat[{\footnotesize$\ratio < \unreachthreshold$}\label{fig:splitpartition:4}]{%
      \includegraphics[width=\splitfigwidth\textwidth]{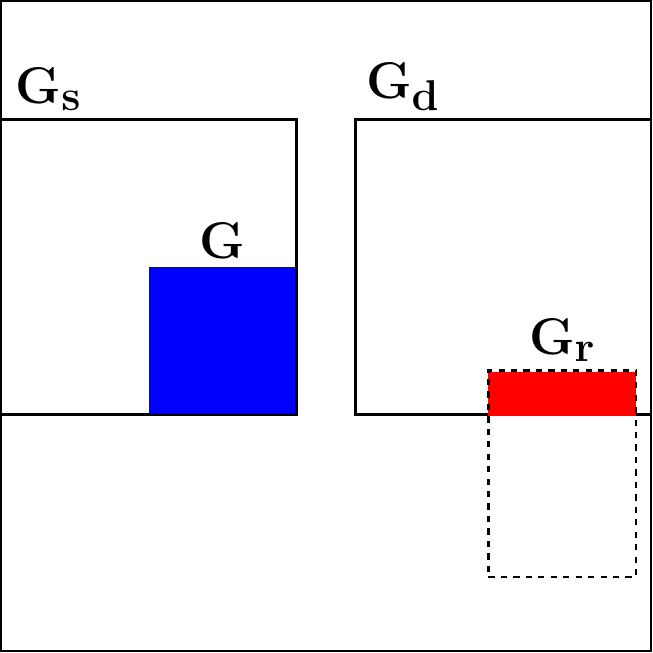}
    }
  \end{center}
  \caption{\textit{SplitSet} 
   splits set $\partition$ in \ref{fig:splitpartition:1}, concludes that $\partition$ reaches $\partition_d$ in \ref{fig:splitpartition:2}, or that $\partition$ does not reach $\partition_d$  in \ref{fig:splitpartition:4}.}
  \label{fig:splitpartition}
\end{figure}

\section{Evaluation}
\label{sec:evaluation}

\subsection{Experimental setup}
We evaluate \acronym with the following research questions:
\begin{inparaenum}
\item[{\bf RQ1)}] Can \acronym learn an interpretable goal space representation that decomposes the task structure?
\item[{\bf RQ2)}] Is the representation effective to achieve better data-efficiency? 
\item[{\bf RQ3)}] Is the representation transferable to new environments?
\end{inparaenum}

\paragraph{Maze Environments}
we evaluate the research questions solving  navigation tasks in a maze. In our experiments, we use a U-shaped maze (Figure~\ref{fig:U-shaped_repr}) and a 4-rooms maze (Figure~\ref{fig:4Rooms_repr}).
The environment's states $[x, y, v_x, v_y]$ are continuous: $(x,y)$ are the agent's position and $v_x, v_y$ are the agent's velocities on the x-axis and y-axis respectively. The agent has 4 actions to control its acceleration: UP/DOWN/LEFT/RIGHT.
%
The agent receives a positive reward only when it reaches the exit position in the maze (i.e., the reward is {\bf sparse}).

Such learning tasks are ideal to evaluate if our algorithm discovers an interpretable representation that decomposes the task (i.e., {\bf RQ1}). In fact, we can intuitively decompose it into subtasks that correspond to moving from one room to the other, and we can compare such representation with the one \acronym learns automatically.
Moreover, in our settings the reward is sparse and not continuous (i.e., the euclidean distance from the exit) as in other settings (e.g., \cite{DBLP:conf/nips/ZhangG0H020}).
Such sparse reward setting is challenging for existing Feudal HRL approaches that, without a proper goal representation, struggle on both levels of the hierarchy.
%
Observe that, despite having a modest continuous state space (i.e., 4 dimensions), the above learning tasks are \emph{non-trivial}: the reward signal is sparse and reaching the exit requires multiple manoeuvres around obstacles. We leave the evaluation of \acronym on higher-dimensional environment as future work.

\paragraph{Baselines}
we compare \acronym with the following combinations of HRL algorithms and representations:

\noindent {\it - Feudal HRL with {\bf Handcrafted} representation:} 
 Similar to \cite{hdqn}, we manually provide a fixed goal representation (Fig. ~\ref{fig:U-shaped_handcrafted_repr}, ~\ref{fig:4Rooms_handcrafted_repr}) that intuitively decomposes the tasks and that {\bf remains unchanged} during training.

\noindent {\it - Feudal HRL with {\bf Concrete} representation:} a deep-learning version of a Feudal HRL algorithm~\cite{feudal} where the high-level agent directly selects {\bf raw states} as goals.

\noindent {\it- Feudal HRL with {\bf LSTM}:} variation of {\bf Concrete} where the high-level agent uses features of a LSTM as input to capture the time dependencies between goals in the task. The LSTM takes as input environment's states and the previous action. 

\noindent {\it - {\bf HIRO}:} this is our implementation of the state-of-art Feudal HRL algorithm HIRO~\cite{nachum2018dataefficient}, which uses raw states as goals but uses a correction mechanism to address non-stationarity.

The high-level agents in GARA and the Handcrafted approach use Q-learning (both the goals and action space are discrete and finite). We use DDPG~\cite{ddpg} to learn high-level policies in Concrete, LSTM, and HIRO, since they directly sample goals from the state space.
To ensure a fair comparison, all the approaches use DQN~\cite{dqn} with the same network architecture for learning the low-level policy.

\paragraph{Experiments}
We run \acronym and the above  algorithms on both the U-shaped and 4-rooms maze.
All the results are obtained over 20 runs of each algorithm (we report the average success rate with its standard deviation).
Further details regarding hyperparameters and additional results can be found in the paper's extended version~\cite{garareport} and the code is available at \url{http://www.lix.polytechnique.fr/Labo/Mehdi.ZADEM/GARA/code.tar.gz}

\subsection{Results}
\begin{figure}[tbh]
\vspace{-0.5cm}
    \centering
    \subfloat[Handcrafted representation \label{fig:U-shaped_handcrafted_repr}]{\includegraphics[width=0.2\columnwidth]{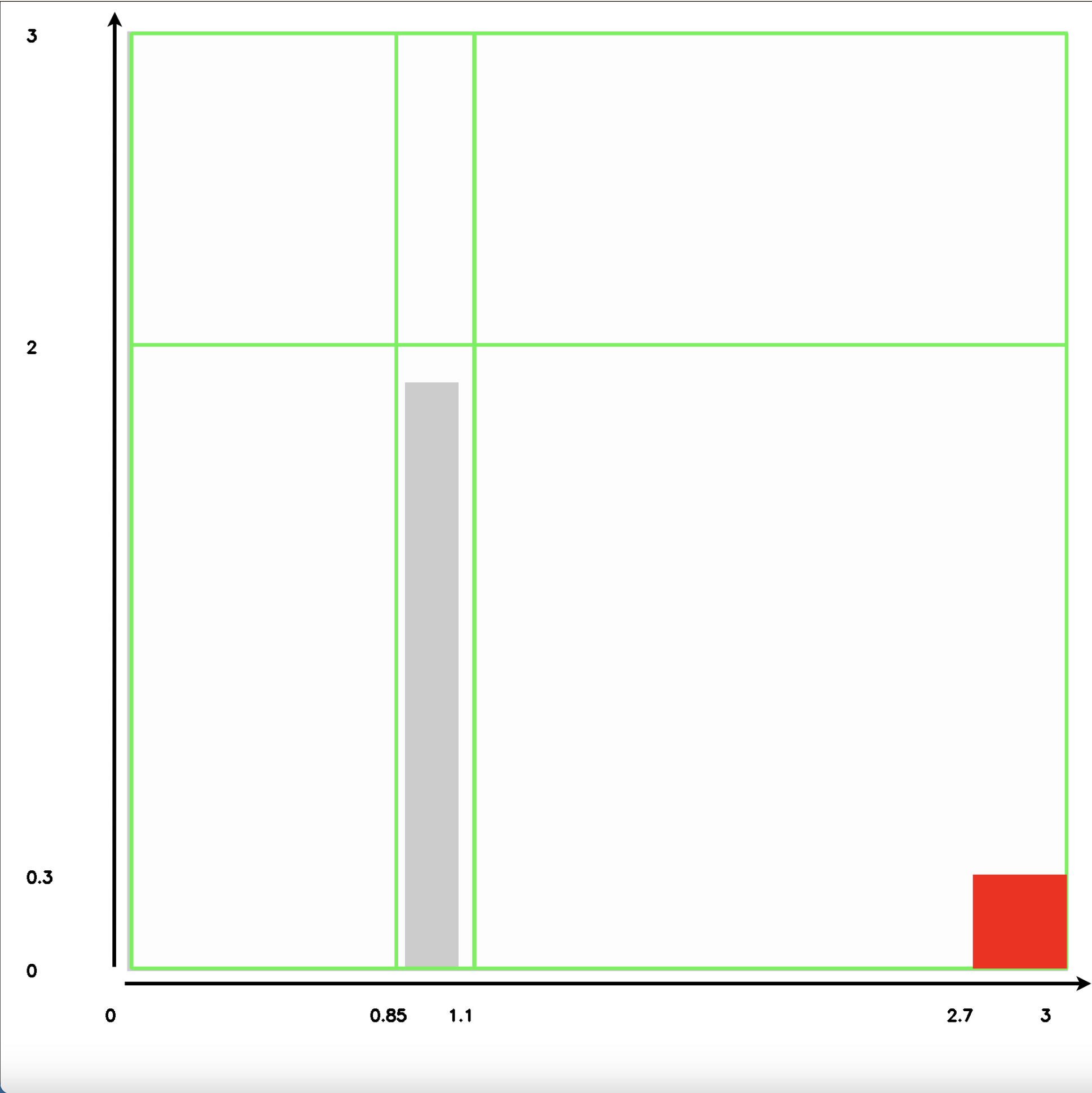}} 
    \hfill
    \subfloat[$\partitions$ by \acronym via clustering after $10^3$ steps \label{fig:U_shaped_repr1}]{\includegraphics[width=0.2\columnwidth]{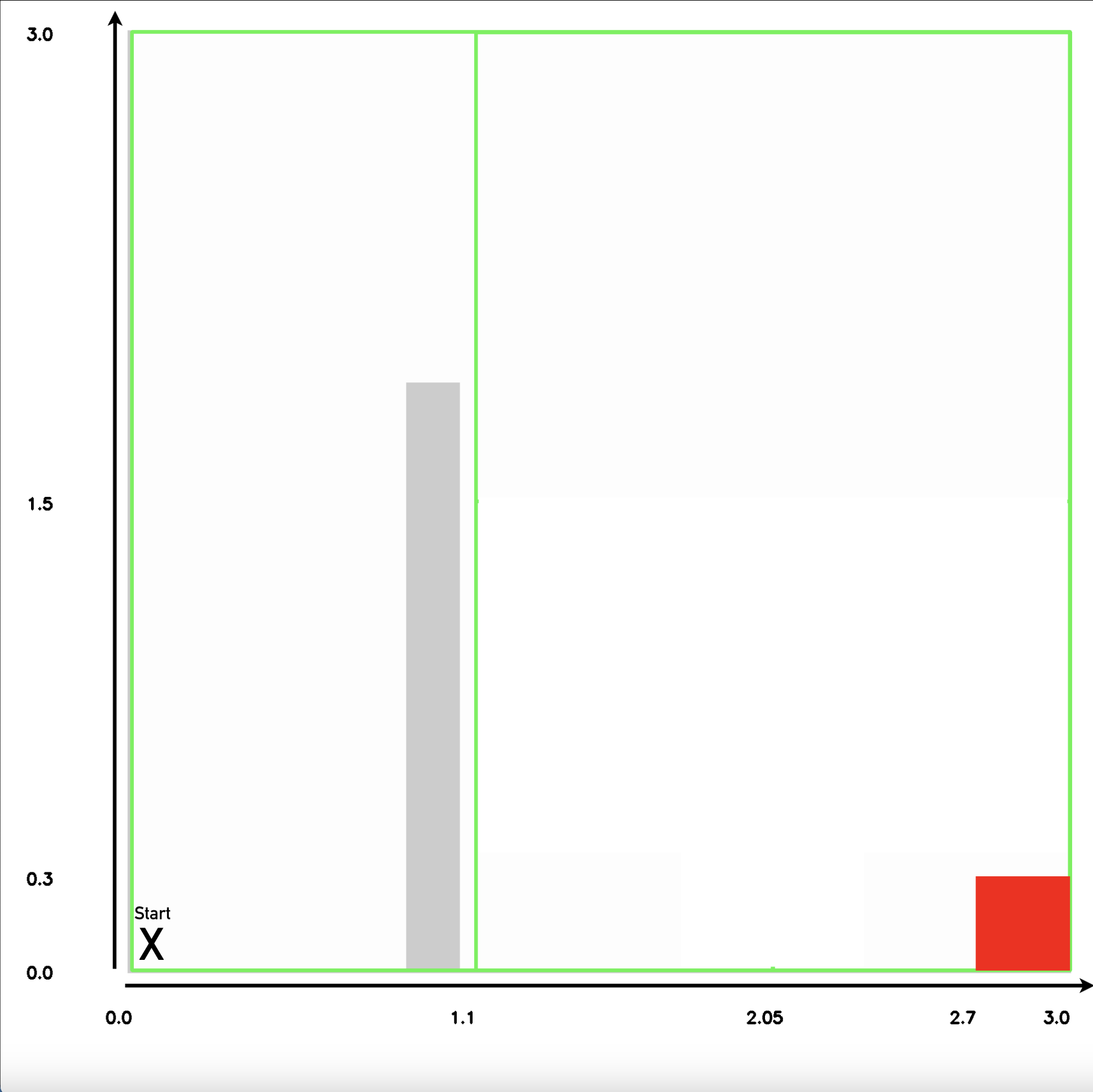}}
    \hfill
    \subfloat[$\partitions$ learned by \acronym after $10^4$ steps \label{fig:U_shaped_repr2}]{\includegraphics[width=0.2\columnwidth]{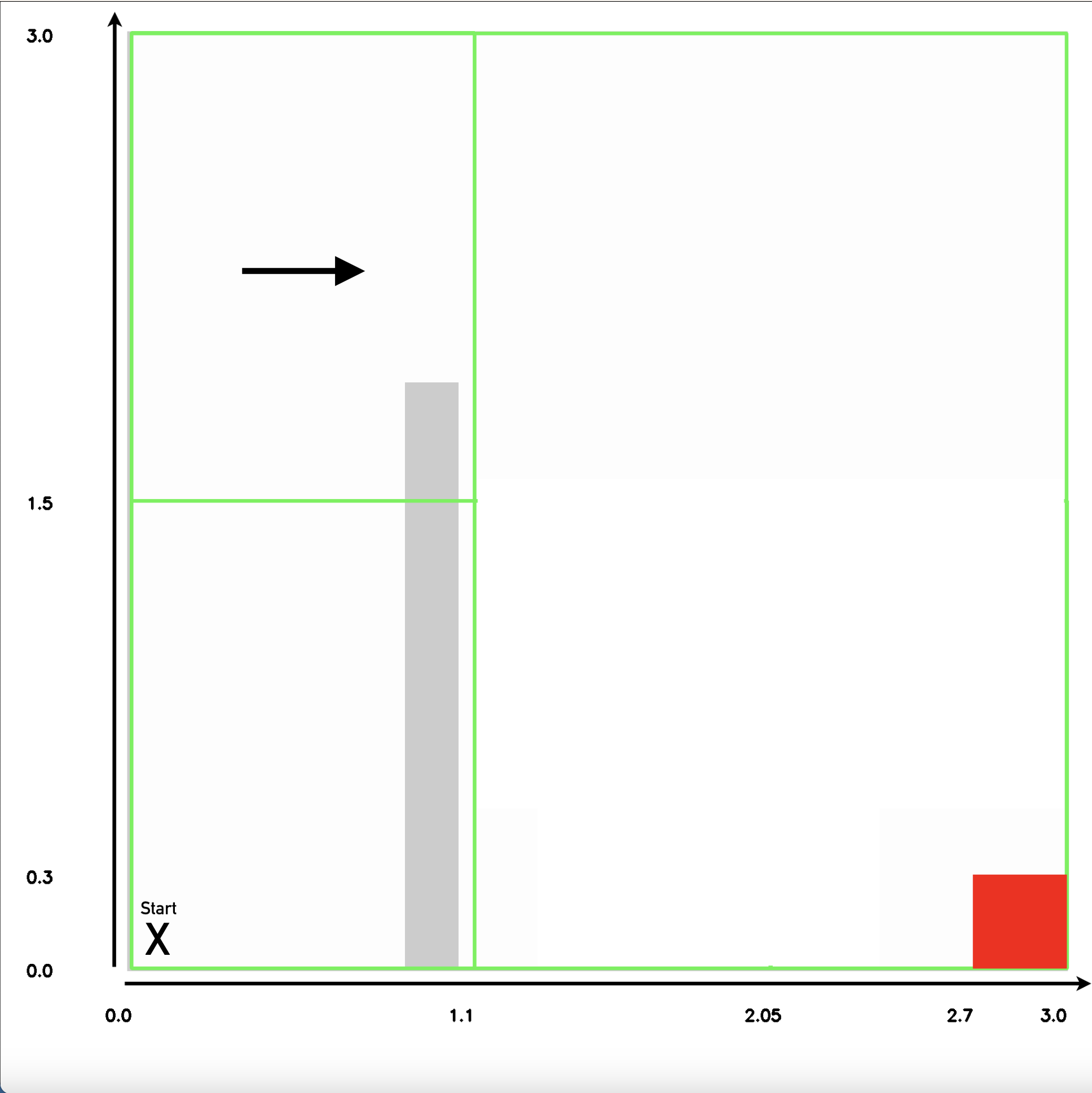}} 
    \hfill
    \subfloat[Final $\partitions$ learned by \acronym \label{fig:U_shaped_repr3}]{\includegraphics[width=0.2\columnwidth]{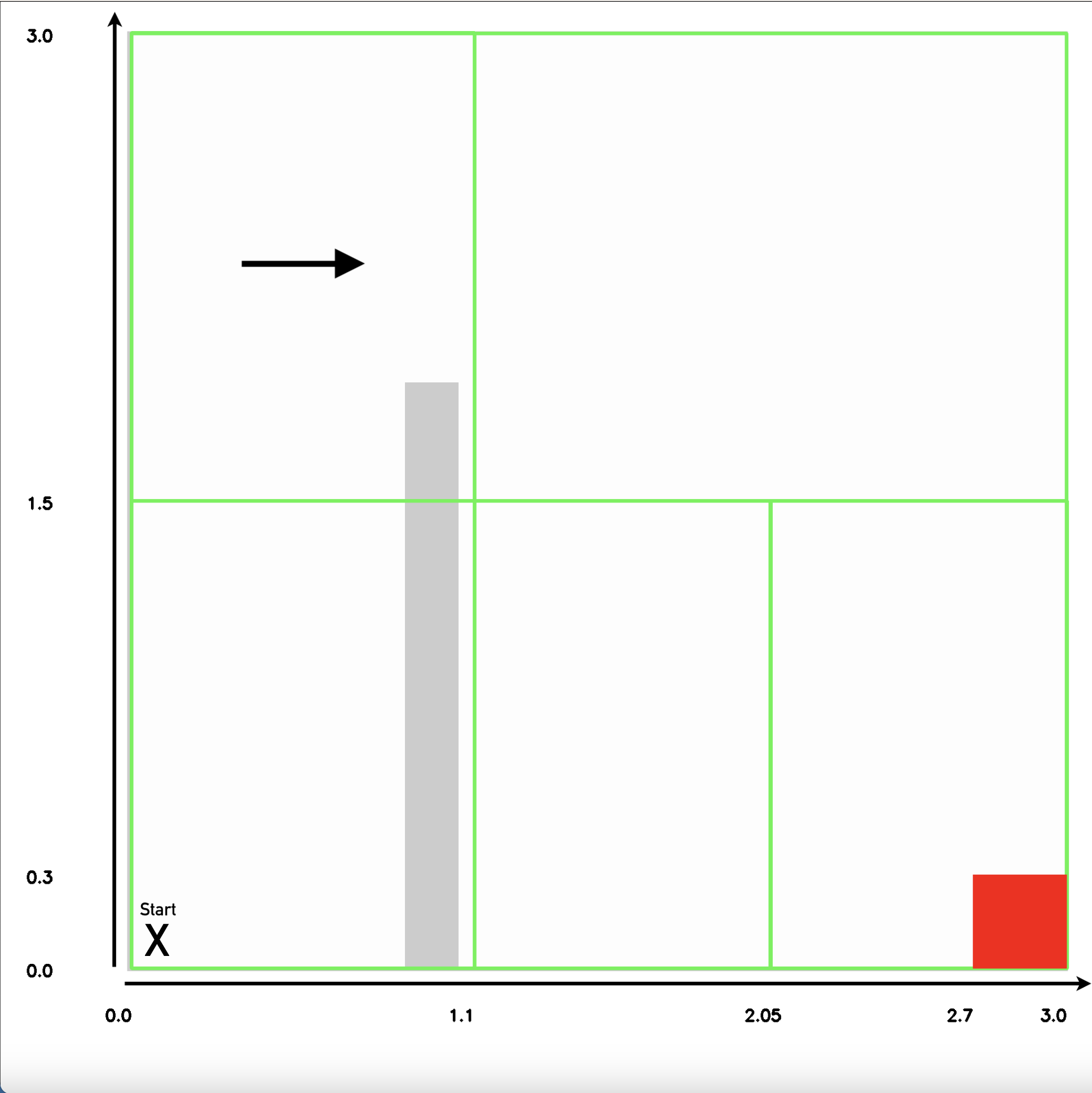}} 
     \caption{Representation of the goal space $\partitions$ in the U-shaped maze for one run of algorithm. The exit is marked in red.
     Green boxes show intervals for $x,y$ and the horizontal and vertical arrows indicate the sign of the velocities $v_x$ and $v_y$, respectively. No arrows indicate there are no split across $v_x$ or $v_y$.}
    \label{fig:U-shaped_repr}
    \vspace{-0.5cm}

\end{figure}

\paragraph*{{\bf RQ1} - representation learning} Focusing first on the learned representation by \acronym, Fig.~\ref{fig:U_shaped_repr1}, Fig.~\ref{fig:U_shaped_repr2}, and Fig.~\ref{fig:U_shaped_repr3} show the evolution of the goal space throughout the learning at $0$, $10^3$, and $3 \times 10^4$ steps (for a randomly selected run of the algorithm).
Starting from an initial partition $\partitions = \{\states\}$, the clustering mechanism in \acronym identifies the left half of the maze as the set of states most explored (Fig.~\ref{fig:U_shaped_repr1}). \acronym then splits the left half of the maze across the y axis and the $v_x$ axis (Fig.~\ref{fig:U_shaped_repr2}) with the reachability analysis mechanism, discovering the region at the top-left corner of the maze with positive velocity $v_x$ (indicated with $\rightarrow$).
Intuitively, such region provides a good starting point to learn policies that efficiently manage to reach the right half of the maze.
In the final partition (Fig.~\ref{fig:U_shaped_repr3}), \acronym refines the right half region, which allows the algorithm to focus on reaching the region containing the exit point.
Our intuition is that such final partition results in easier to reach goals, prompting the agent to select successful behaviours (i.e., the high-level agent samples goals that are reachable from the current state and that can, in turn, lead to a goal containing the exit). 
In the final representation, only the top-left region is split across a velocity variable. This is justified by the complexity of the maneuvers where a smaller set of states (where $v_x > 0$) reach the right side.  
%
Overall, Fig.~\ref{fig:U-shaped_repr} shows that \acronym learns an interpretable representation from data collected during the HRL exploration.

\begin{figure}[tbh]
\vspace{-0.2cm}
    \centering
    \includegraphics[width=1.0\columnwidth]{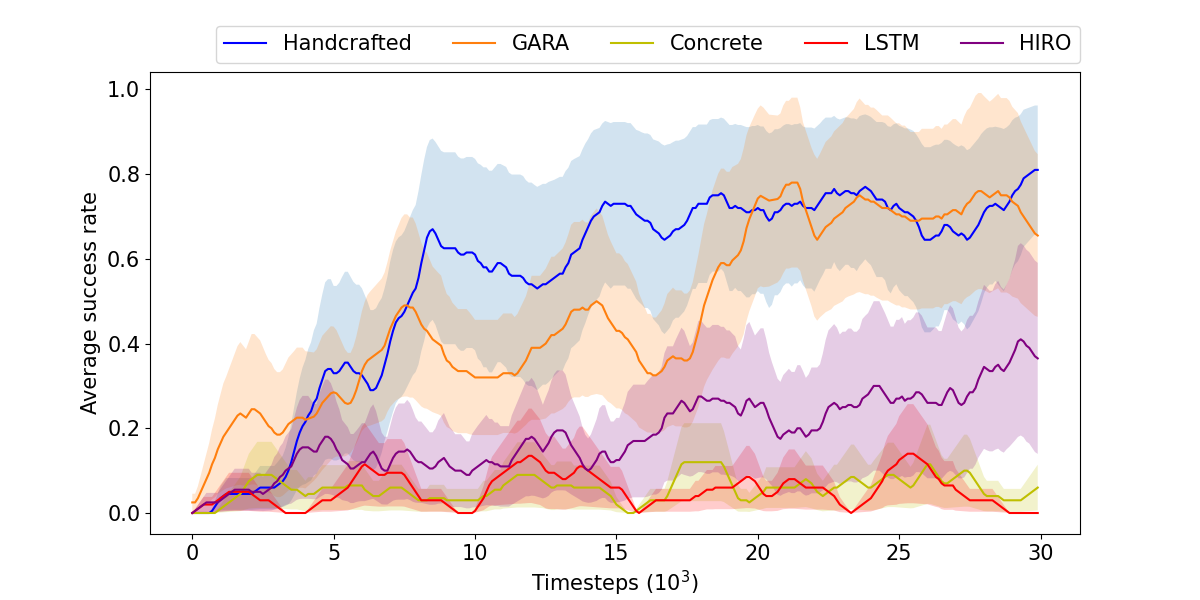}
    \caption{Average success rate on the U-shaped Maze (over 20 runs).}
    \label{fig:U_shaped}
\vspace{-0.2cm}
\end{figure}

\paragraph*{{\bf RQ2} - data efficiency} Fig. \ref{fig:U_shaped} shows that \acronym learns a successful hierarchical policy with a performance approaching the handcrafted representation, whereas all the other approaches, including HIRO, cannot learn to solve the task within the same time frame. We attribute this performance to the better sample efficiency due to the learned abstraction, as the agents successfully decompose the task into simple-to-achieve goals. We note that the handcrafted approach starts improving after \acronym since initially it must explore more regions.

\begin{figure}[tbh]
\vspace{-0.5cm}
    \centering
    \subfloat[Handcrafted representation \label{fig:4Rooms_handcrafted_repr}]{\includegraphics[width=0.2\columnwidth]{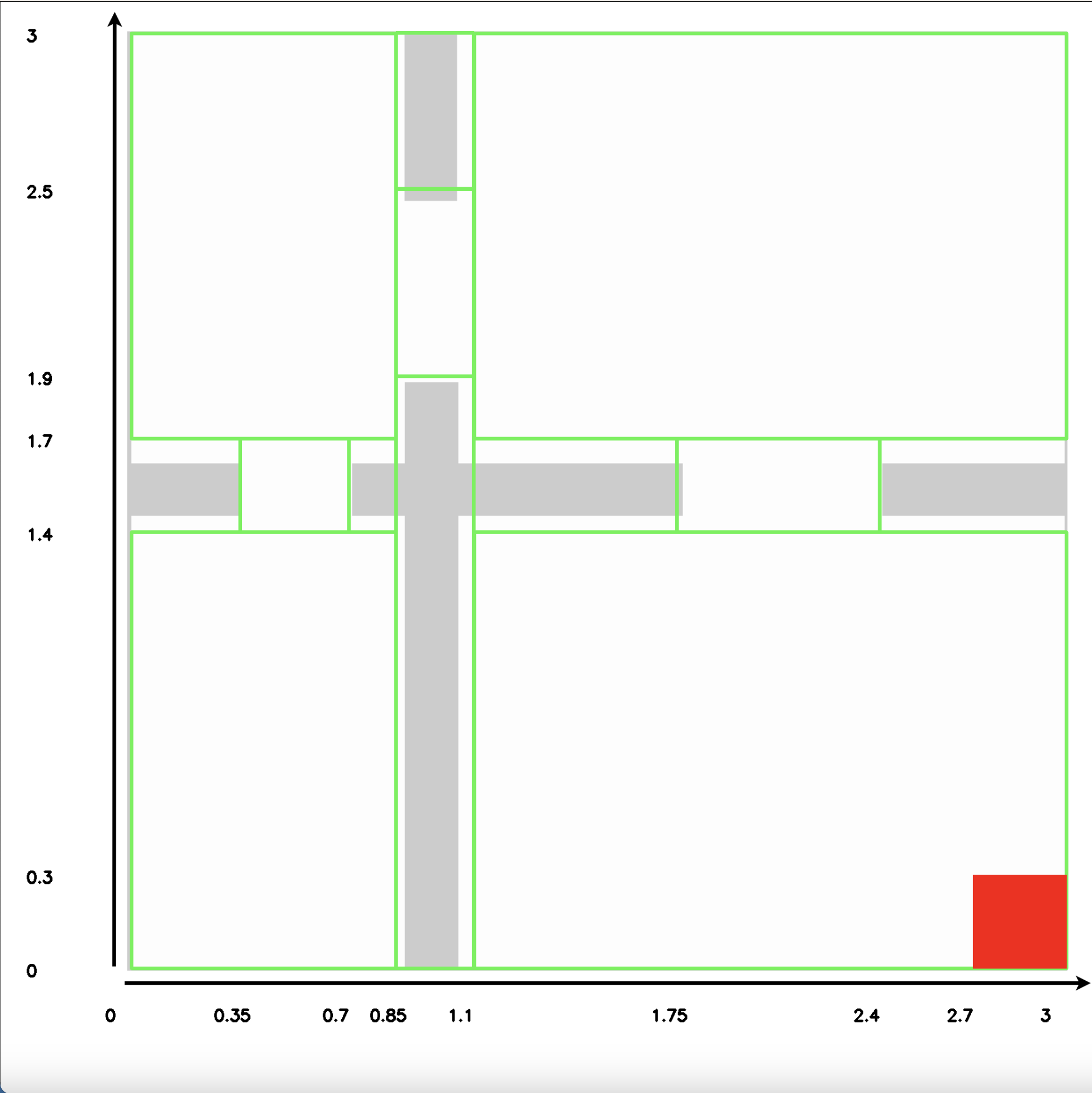}} 
    \hfill
    \subfloat[Transferred $\partitions$ for \acronym-T \label{fig:4Rooms_repr1}]{\includegraphics[width=0.2\columnwidth]{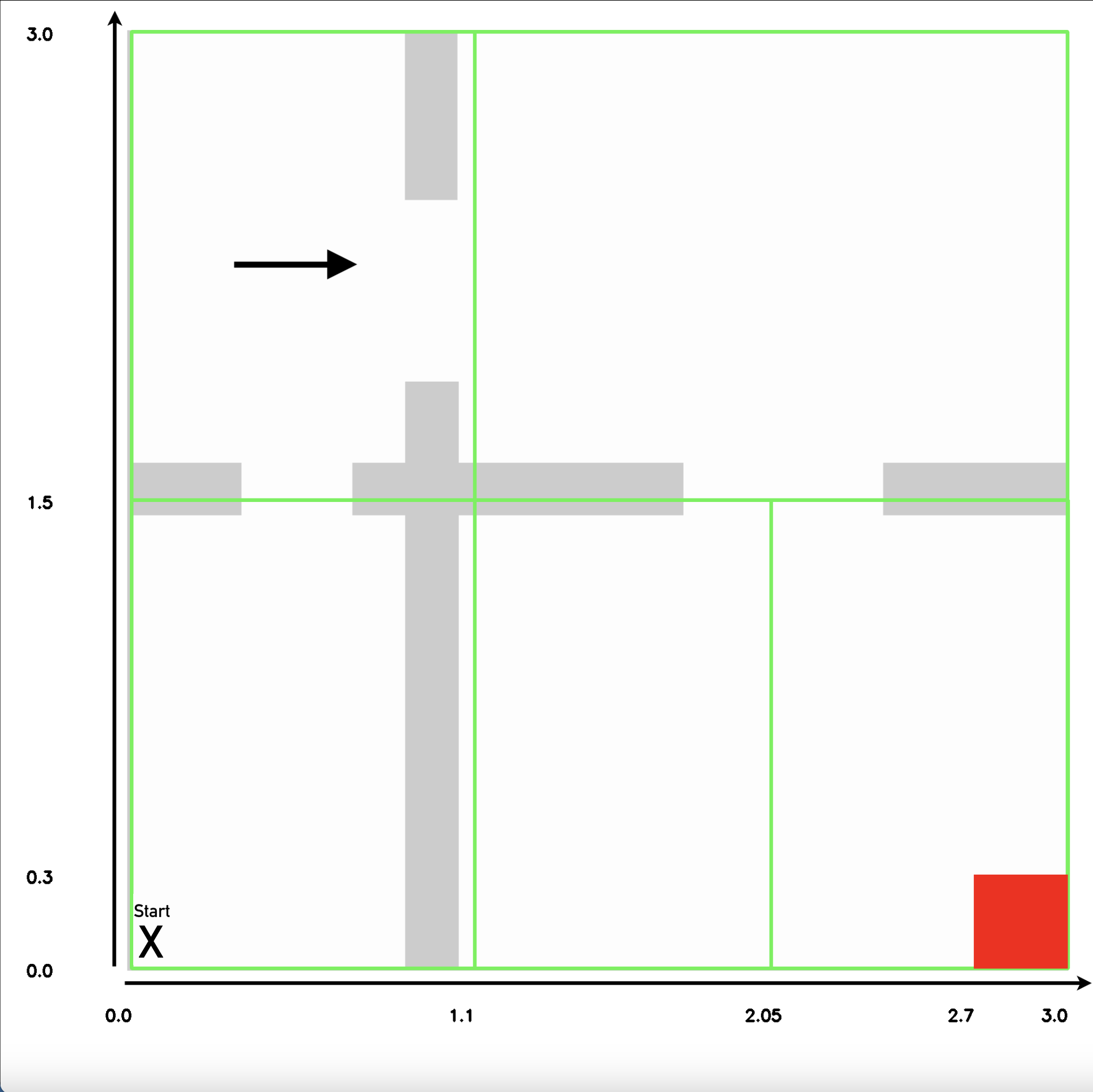}}
    \hfill
    \subfloat[$\partitions$ learned by \acronym-T after $10^4$ steps \label{fig:4Rooms_repr2}]{\includegraphics[width=0.2\columnwidth]{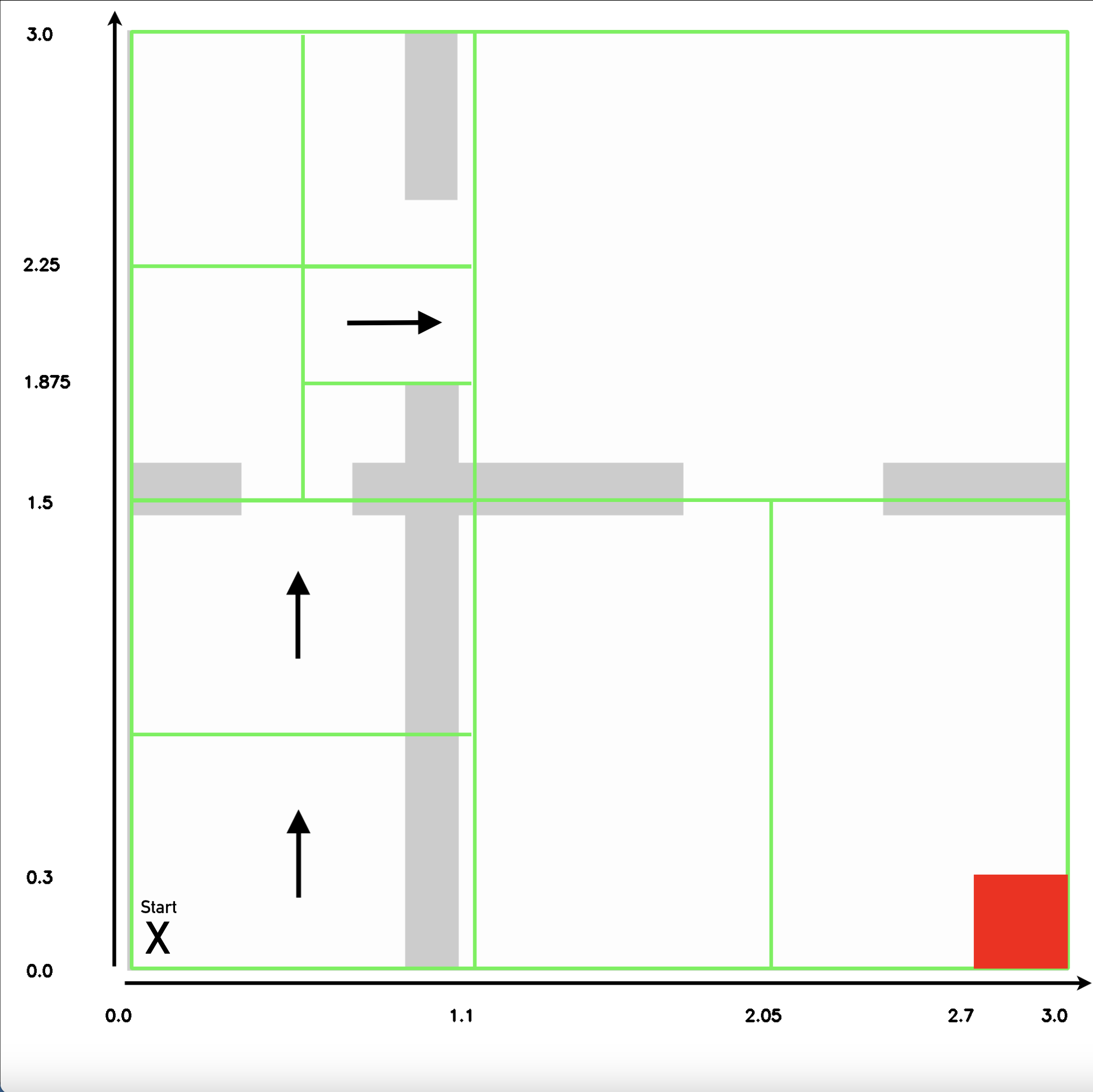}} 
    \hfill
    \subfloat[Final $\partitions$ learned by \acronym-T \label{fig:4Rooms_repr3}]{\includegraphics[width=0.2\columnwidth]{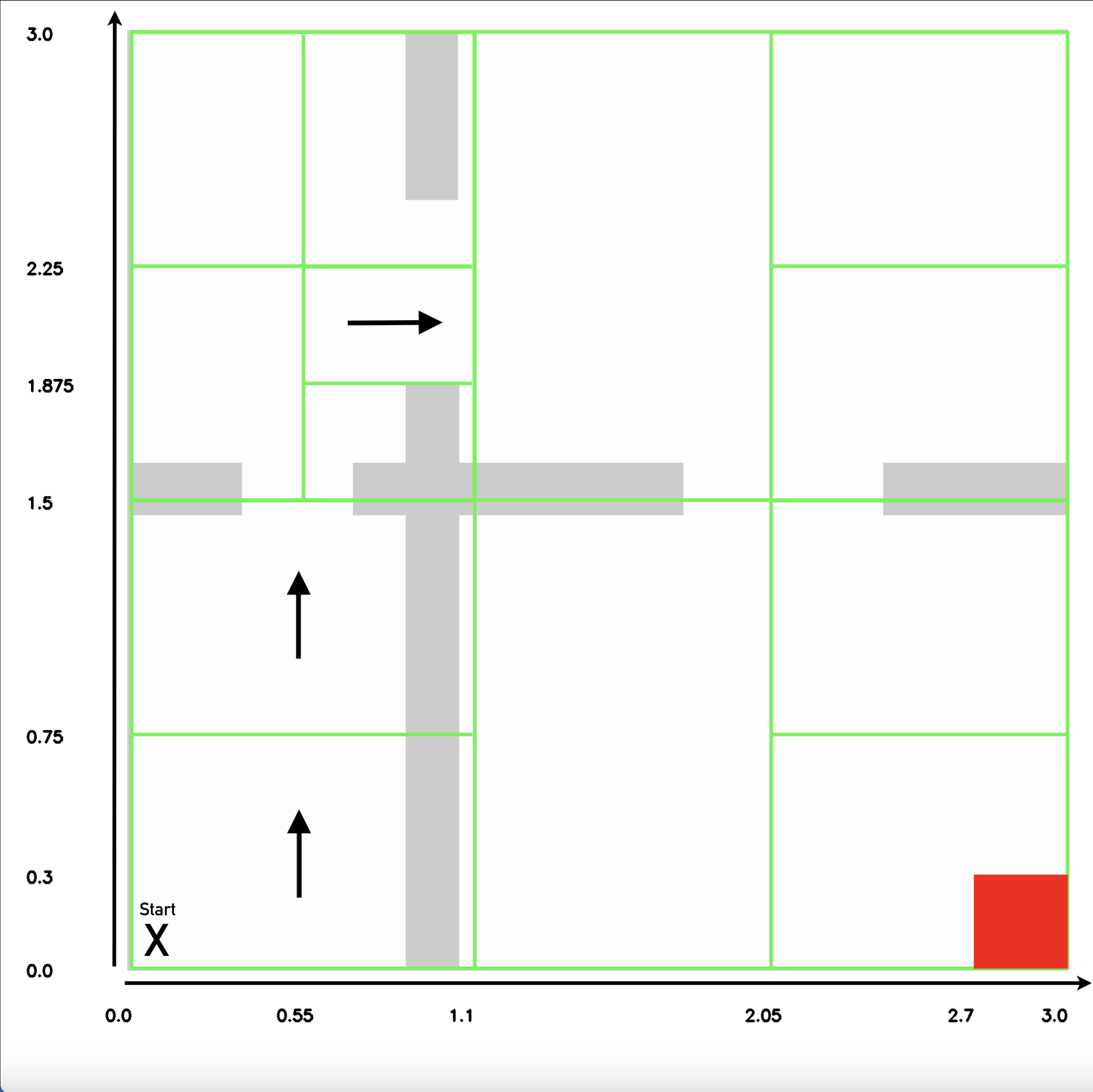}} 
\vspace{-0.2cm}
     \caption{Goal space representations for the 4-Rooms Maze}
    \label{fig:4Rooms_repr}
\vspace{-0.5cm}
\end{figure}

\paragraph*{{\bf RQ3} - transfer learning}
We examine the transferability of the goal space  representation when we change the environment's configuration from the U-shaped to the more challenging 4-rooms one.
For \acronym, we transfer the representation learned from the first task (Fig. \ref{fig:U_shaped_repr3}), while for Feudal HRL with Concrete representation, LSTM, and HIRO we transfer the weights of the high-level agent's neural networks. We also provide a new representation to the handcrafted approach (Fig.~\ref{fig:4Rooms_handcrafted_repr}).

Fig. \ref{fig:4Rooms_repr2} shows that GARA takes advantage of the transferred representation (Fig. \ref{fig:4Rooms_repr1}), further refining the left side of the maze to better respect its geometry and the smaller openings. Later in the training (Fig.~\ref{fig:4Rooms_repr3}), the right side of the maze is refined to adapt to the new walls, consequently devising different policies in each of these sub-regions. This allows the agent to clearly formulate a goal identifying the openings to go past the obstacles and reach the exit.

\begin{figure}[tbh]
    \centering
    \includegraphics[width=1.0\columnwidth]{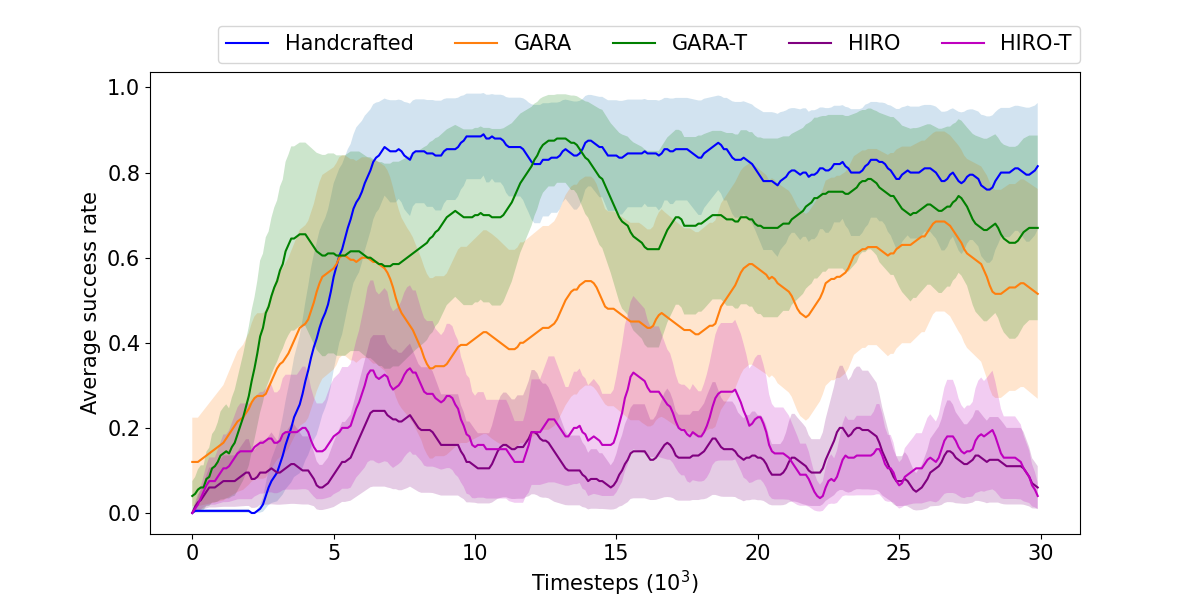}
    \caption{Average success rate on the 4-Rooms Maze (over 20 runs). -T refers to the transferred versions of the algorithms.}
    \label{fig:4rooms}
\vspace{-0.5cm}
\end{figure}

We report the performance of the approaches learning from scratch and with transfer learning in Fig.~\ref{fig:4rooms}. To keep the plot readable, we omit the concrete representation and LSTM approaches that always fail to learn a successful policy.
From Fig. \ref{fig:4rooms}, we first observe that the performance of HIRO, which uses raw states as goals, do not improve with the transfer.
Instead, we observe that \acronym with transfer learning performs better than \acronym from-scratch, reaching a reward close to the manual representation. Thus, \acronym successfully leverages the learned representation and adapts it to solve a more complex task.

\section{Related Works}

Learning and acting over different space and time resolutions simultaneously is a key challenge in tasks involving long-range planning. Sutton et al.~\cite{Sutton1999AI} proposed the options framework for a temporal abstraction. Recent approaches focus on state representation. Vezhnevets et al.~\cite{DBLP:journals/corr/VezhnevetsOSHJS17} proposes a feudal HRL~\cite{feudal} that learns a low-dimensional latent state space for subgoal selection, which still fails to capture critical information (e.g. temporal dependencies) to solving the task as argued in \cite{nachum2018nearoptimal}. 
\cite{DBLP:conf/nips/ZhangG0H020,DBLP:conf/iclr/GhoshGL19,DBLP:conf/iclr/SavinovRVMPLG19,DBLP:conf/nips/EysenbachSL19,li2021learning} propose subgoal sampling methods that are aware of the environment dynamics. Closer to our idea of analysing reachability, Zhang et al. \cite{DBLP:conf/nips/ZhangG0H020} propose to learn a distance-based adjacency relation between subgoals in a predefined space and use it to sample reachable subgoals. Ghosh et al. \cite{DBLP:conf/iclr/GhoshGL19} propose to learn state embeddings using the action distribution of a goal-conditioned policy but this requires already successful policies. Savinov et al. \cite{DBLP:conf/iclr/SavinovRVMPLG19} propose a supervised learning approach for estimating reachabilty relations between states. Li et al. \cite{li2021learning} propose to learn a stable state embedding along with an exploration policy that aims for novel reachable subgoals. 
While these representations remain in the continuous domain, \acronym's abstraction constitutes a symbolic representation of a multi-dimensional continuous state space. It thus bridges the continuous world with a symbolic description, opening the door to methods such as planning, symbolic reasoning or interpretability. Indeed, Kulkarni et al. \cite{hdqn} also resort to a discrete, manual, object-based subgoal representations where goals are modeled as sets of states. In comparison, our approach could find an emerging abstraction by learning online.
Alternatively, Nachum et al. \cite{nachum2018dataefficient} proposed an approach that relies on sampling subgoals directly from the state space. However, this unconstrained selection can result in learning inefficiency and suboptimal policies.

Symbolic methods in RL were studied in works like \cite{DBLP:journals/corr/GarneloAS16}. Lyu et al. \cite{DBLP:conf/aaai/LyuYLG19} incorporate symbolic reasoning to function as planning for RL agents. However, this requires manual symbolic domains to abstract the states, as well as complete descriptions for the learning objective.
Other approaches alleviate this by learning the planning instruction on a set of given abstract states \cite{DBLP:conf/aips/IllanesYIM20,DBLP:journals/corr/abs-2103-08228,Hasanbeig:2021:DAS}. Eventually, the learned high-level behaviour of the agent produces faster learning and can be expressed as a logical formula that allows interpretability and transfer learning. Our work achieves these results without relying on given abstractions.

A few approaches have tacked emergence of representations for hierarchical tasks in open-ended learning. IM-PB \cite{Duminy2018IIRC} learns a recursive task decomposition for  hierarchical tasks of various complexities with intrinsic motivation. CURIOUS \cite{DBLP:conf/icml/ColasOSFC19} selects goals from a set of pre-defined modules. \cite{DBLP:conf/nips/ColasKLDMDO20} uses language instructions as an abstract goal space, and can imagine new goals by language compositionality. In \cite{ugur_emergent_structuring}, nested modules of different complexity levels emerge from predefined low-level features and a list of predefined affordances. 
\acronym allows for more flexible goal definitions without specific instructions or behaviour descriptions.

\section{Conclusion}
\label{sec:conclusion}
In this paper, we tackled the problem of learning from self-exploration data, a symbolic goal representation of a continuous state space  that efficiently captures the task structure, is transferable and interpretable.
We first proposed a representation abstracting into the same goal all the states that reach the same goal with the same low-level policy. This representation as a set of states is interpretable and captures the relation between high- and low-level policies.
We then introduced \acronym, a HRL algorithm that  discovers such representation online, while also learning the agents' policies.
\acronym's representation emerges dynamically from the data acquired during exploration. 
Technically, \acronym refines the representation from the reachability relations approximated with a neural network and studied with set-based reachability analysis.  
Experimental results show that \acronym gradually learns a data-efficient goal representation that is \emph{necessary} to solve tasks on complex continuous environments with sparse rewards. Moreover, the experiments show that \acronym's representation is interpretable and transferable.

In the future, we plan to learn a representation in higher-dimensional environments, where learning low-level policies is harder and an interval set representation is not sufficient.


\bibliographystyle{IEEEtran}
\bibliography{IEEEabrv,aaai23.short.bib}

\newpage

\end{document}